\documentclass[submission,copyright,creativecommons]{eptcs}
\providecommand{\event}{ML4P 2018} 
\usepackage{underscore}           
\usepackage{algpseudocode}
\usepackage{amsmath}
\usepackage{color}
\usepackage{doi}
\usepackage{enumitem}
\usepackage{etoolbox}
\usepackage{fancyvrb}
\usepackage{float}
\usepackage{multicol}
\usepackage{pbox}
\usepackage{romannum}
\usepackage{tabularx}
\usepackage{subfig}
\usepackage{graphicx}
\usepackage{caption}
\usepackage{listings}
\usepackage[toc,page]{appendix}
\usepackage{hyperref}
\definecolor{codegreen}{rgb}{0,0.6,0}
\definecolor{codegray}{rgb}{0.5,0.5,0.5}
\definecolor{codepurple}{rgb}{0.58,0,0.82}
\definecolor{backcolour}{rgb}{0.95,0.95,0.92}
\lstdefinestyle{mystyle}{
    commentstyle=\color{codegreen},
    keywordstyle=\color{magenta},
    numberstyle=\tiny\color{codegray},
    stringstyle=\color{codepurple},
    basicstyle=\footnotesize,
    breakatwhitespace=false,         
    breaklines=true,                 
    captionpos=b,                    
    keepspaces=true,                 
    showspaces=false,                
    showstringspaces=false,
    showtabs=false,                  
    tabsize=2,
    frame=single
}
 
\title{Splitting source code identifiers using\\ Bidirectional LSTM Recurrent Neural Network}
\author{Vadim Markovtsev\email{vadim@sourced.tech} \and
Waren Long\email{waren@sourced.tech} \and
Egor Bulychev\email{egor@sourced.tech}  \and
Romain Keramitas\email{romain@sourced.tech}  \and
Konstantin Slavnov\email{konstantin@sourced.tech}  \and
Gabor Markowski\email{gabor@sourced.tech}}
\def\titlerunning{Splitting source code identifiers using Bidirectional LSTM Recurrent Neural Network}
\def\authorrunning{Markovtsev et al.}
\begin{document}
\maketitle
\begin{center}
\vspace*{-1.5em}
source\{d\}, Madrid, Spain
\vspace*{1em}
\end{center}

\def \pgaSize {182,014 }
\def \identifierLengthLimit {40 }
\def \charStatModelDepth {11}

\def \bilstmLayerSize {256}

\def \datasetRecords {62.0 }
\def \datasetRecordsAfterManualFiltering {49.2 } 
\def \datasetRecordsAfterHeuristicSplit {36.1 } 
\def \datasetRecordsAfterLengthThreshold {34.9 }

\def \ratioMultipartTokens {97}
\def \ratioNotSplittableByHeuristics {7.5}
\def \ratioFurtherSplittableByBestModelAfterHeuristics {15}
\def \ratioCoreVocabularyReduced {64}
\def \ratioCoreVocabularyTimes {2}
\def \ratioCoreVocabularyBefore {2,940,710}
\def \ratioCoreVocabularyAfter {1,065,153}

\def \bilstmPrecision {95}
\def \bilstmRecall {96}
\def \bilstmIdentifiersToHalfError {14} 

\def \bilstmHoursToRunEpochs {14 }
\def \bilstmNumEpochs {10 }

\begin{abstract}

Programmers make rich use of natural language in the source code they write through identifiers and comments. Source code identifiers are selected from a pool of tokens which are strongly related to the meaning, naming conventions, and context. These tokens are often combined to produce more precise and obvious designations. Such multi-part identifiers count for \ratioMultipartTokens\% of all naming tokens in the \emph{Public Git Archive} - the largest dataset of Git repositories to date. We introduce a bidirectional LSTM recurrent neural network to detect subtokens in source code identifiers. We trained that network on \datasetRecordsAfterLengthThreshold million distinct splittable identifiers collected from \pgaSize open source projects in Public Git Archive, and show that it outperforms several other machine learning models. The proposed network can be used to improve the upstream models which are based on source code identifiers, as well as improving developer experience allowing writing code without switching the keyboard case.

\end{abstract}
 
\section{Introduction}

The descriptiveness of source code identifiers is critical for readability and maintainability \cite{1631100}. This property is hard to ensure by using exclusively single words. Therefore it is common practice to concatenate several multiple words into a single identifier. Whitespace characters in identifiers are forbidden by most programming languages, so there are naming conventions \cite{namingconventions} like \emph{CamelCase} or \emph{snake_case} which specify the concatenation rules. It is possible to apply simple heuristics, backtrack those rules and restore the original words from identifiers.
For example, \texttt{FooBar} or \texttt{foo_bar} are trivially disassembled into \texttt{foo} and \texttt{bar}. However, if there is a compound identifier consisting of
only lowercase or only uppercase characters, splitting requires domain knowledge and cannot be easily performed.

According to our estimations, up to \ratioNotSplittableByHeuristics\% of the identifiers are not splittable by style driven heuristics; among the rest, \ratioFurtherSplittableByBestModelAfterHeuristics\% contain further splittable parts after heuristics. This leads to bigger vocabulary sizes, worse performance, and reduced quality of upstream investigation in the areas of source code analysis and Machine Learning on Source Code (MLonCode). A deep learning-based parser, capable of learning to tokenize identifiers from many training examples, can enhance the quality of research in topics like identifier embeddings \cite{Nguyen:2017:EAE:3097368.3097421}, deduplication \cite{Lopes:2017:DMC:3152284.3133908}, topic modeling \cite{Markovtsev:2017}, and naming suggestions \cite{Allamanis:2015:ACM:2786849}.

The main contributions of this paper are:
\setlist{nolistsep}
\begin{itemize}[noitemsep]
\item We built the biggest dataset of \datasetRecordsAfterManualFiltering million source code identifiers extracted from \emph{Public Git Archive} \cite{pga} - The \pgaSize most popular GitHub repositories.
\item We are the first to apply a recurrent neural network to split "unsplittable" identifiers. We show that the character-level bidirectional LSTM type of recurrent neural network (RNN) performs better than the character-level bidirectional GRU, character-level convolutional neural network, the gradient boosted decision tree, the statistical dynamic programming model, and the unsmoothed maximum likelihood character-level model.
\end{itemize}

\section{Identifier extraction}

This section describes the source code identifier extraction pipeline which was used to generate the train dataset from the \emph{Public Git Archive}.

We processed each Git repository with the \textbf{ source\{d\} engine} \cite{Engine} to determine the main branch and its head revision, took the files from that revision, and identifed the programming languages
of those files. We extracted identifiers from files according to the identified language with
\textbf{babelfish} \cite{bblfsh} and \textbf{pygments} \cite{pygments}. Babelfish is a self-hosted server for universal source code parsing, it converts code files into Universal Abstract Syntax Trees. We fall back to Pygments for those languages which are not supported yet by Babelfish. Pygments was developed to highlight source code and uses regular expressions, however it makes mistakes and introduces noise.

We obtained \datasetRecords million identifiers after removing duplicates. This number reduced to \datasetRecordsAfterManualFiltering after manual rule-based filtering of noisy output from Pygments. We then split the identifiers according to the common naming conventions. For example \texttt{FooBarBaz} becomes \texttt{foo bar baz}, and \texttt{method_base} turns into \texttt{method base}. The listing of the function code which implements the heuristics is provided in appendix \ref{tokenparser}. We left only those which consisted of more than one part and obtained \datasetRecordsAfterHeuristicSplit million distinct subtoken sequences. Some identifiers aliased to the same subtoken sequence. The distribution of identifier lengths had a long tail as seen on Fig. \ref{distriblen}, so we put a threshold of maximum identifier length to \identifierLengthLimit characters. The length threshold further reduced the dataset to \datasetRecordsAfterLengthThreshold million unique subtoken sequences. All the models we trained used as input the lowercase strings created by merging the subtoken sequences together and the corresponding indices of subtoken boundaries. Figure \ref{distribtoken} depicts the head of the frequency distribution of the subtokens.

The raw dataset of \datasetRecordsAfterManualFiltering million identifiers is available for download on GitHub \cite{dataset}. Currently available datasets of source code identifiers contain less than a million entities and focus on particular programming languages, such as Java \cite{Butler:2013:IIN:2487085.2487159}.

\newcounter{tablecounter}
\newcounter{figurecounter}

\begin{figure}[H]
\setcounter{figure}{\value{figurecounter}}
\centering
\begin{minipage}{.5\textwidth}
\centering
\vspace*{0.7em}
\includegraphics[width=0.95\linewidth]{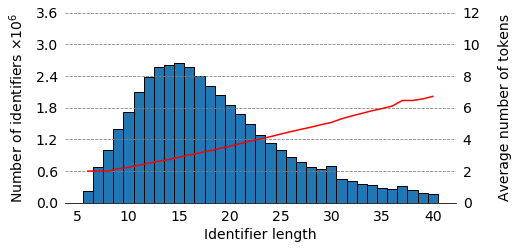}
\vspace*{1.1em}
\captionof{figure}{Distribution of identifier lengths}
\refstepcounter{figurecounter}
\label{distriblen}
\end{minipage}%
\begin{minipage}{.5\textwidth}
\centering
\includegraphics[width=0.95\linewidth]{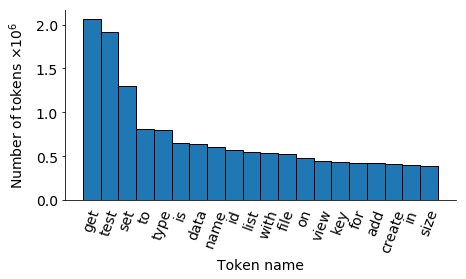}
\vspace*{-0.5em}
\captionof{figure}{Distribution of most frequent subtokens}
\refstepcounter{figurecounter}
\centering
\captionsetup{justification=centering}
\label{distribtoken}
\end{minipage}
\end{figure}

\section{Baselines}

We now describe the models to which we compare our character-level bidirectional LSTM recurrent neural network.

\paragraph{Maximum likelihood character-level model}
Probabilistic maximum likelihood language models (ML LM) are typical for Natural Language Processing \cite{Nguyen:2013:SSL:2491411.2491458}. Given the sequence of characters $a_1, \dots, a_M$ representing an identifier, we evaluate for each character $a_{i+1}$ the probability
that the subsequence $a_1, \dots, a_i$ is a prefix, and pick the prefix that maximizes that probability, assuming  $P(a_1, \dots, a_i)= \prod_{k=1}^{i} P(a_k \mid a_{k-1}, \dots, a_1)$. We repeat this procedure from the character following the chosen prefix. In the case of prefixes for which we have no prior knowledge we slide the root forward until the match is found. Similarly to \textit{n}-gram models \cite{Hindle:2012:NS:233722}, our character-level LM makes the \emph{Markov assumption} that the sequence of characters is a memoryless stochastic process, so we assert that $P(a_i \mid a_{i-1}, \dots,a_1) = P(a_i \mid a_{i-1}, \dots,a_{i-n+1})$. We estimate these conditional probabilities using maximum likelihood \cite{Manning:1999:FSN:311445}. We trained two unsmoothed models independently, corresponding to forward and backward reading direction. Finally, we combined them via the logical conjunction and disjunction. The tree depth was \charStatModelDepth\ due to the technical limitations - bigger depths require too much operating memory. The implementation was \textbf{CharStatModel} \cite{CharStatModel}.

\paragraph{Dynamic programming}
Inspired by the dynamic programming approach to splitting words \cite{Koehn:2003:EMC:1067807.1067833}, we implemented the similar solution based on word frequencies. By making the hypothesis that the words are independent from one another, we can model the probability of a sequence of words using frequencies computed on a corpus. We trained on the generic Wikipedia corpus and on the unique subtokens in our identifier dataset, either assuming Zipf prior or the posterior. Our implementation was based on \textbf{wordninja} \cite{wordninja}.

The main limitation of the statistical approaches is their inability to predict out of vocabulary words, especially method and class names which represent the substantial portion of identifiers in the validation set. The only way to compensate this drawback is to increase the length of the context on which we compute priors, simultaneously worsening the data sparsity problem \cite{Allamanis:2015:ACM:2786849} and increasing the time and memory requirements.

\paragraph{Gradient boosting on decision trees}
We trained the gradient boosting on decision trees (GBDT)) using \textbf{XGBoost} \cite{xgboost}. The tree input was a 10-character window with "a"-aligned ASCII codes instead of one-hot encoding. We didn't choose a larger window to avoid introducing noise, given bulk of our identifiers were shorter then the \identifierLengthLimit character limit. The windows were centered at each split point and we also generated 80\% negative samples at random non-split positions. The maximum tree depth was 30, the number of boosting trees was 50.

\paragraph{Character-level Convolutional Neural Network}
We stacked 3 Inception layers \cite{inception}, with 1-dimensional ReLU kernels spanning over 2, 4, 8, and 16 one-hot encoded characters, and 32 dimensionality reducing ReLU kernels of size 1. Thus the output of each layer was shaped \identifierLengthLimit by 32. The last layer was connected to the time-distributed dense layer with sigmoid activation and binary labels. There was no regularization as the dataset size was big enough and we used RMSProp optimizer \cite{rmsprop}.

\section{Character-level bidirectional recurrent neural network}

Character-level bidirectional recurrent neural networks (BiRNNs) \cite{Schuster:1997:BRN:2198065.2205129} are a family of models that combine two recurrent networks moving through each character in a sequence in the opposite directions and starting from the opposite sides. BiRNNs are an effective solution for sequence modeling, so we tried them for the splitting task. Given that the tokens may be long, we chose LSTM \cite{phd/de/Graves2008} over vanilla RNNs to overcome the vanishing gradients problem. Besides, we compared LSTM to GRU \cite{gru} as GRU was shown to perform with similar quality but are faster to train.

\begin{figure}[H]
\setcounter{figure}{\value{figurecounter}}
\refstepcounter{figurecounter}
\vspace*{-1em}
\centering
\includegraphics[scale=0.18]{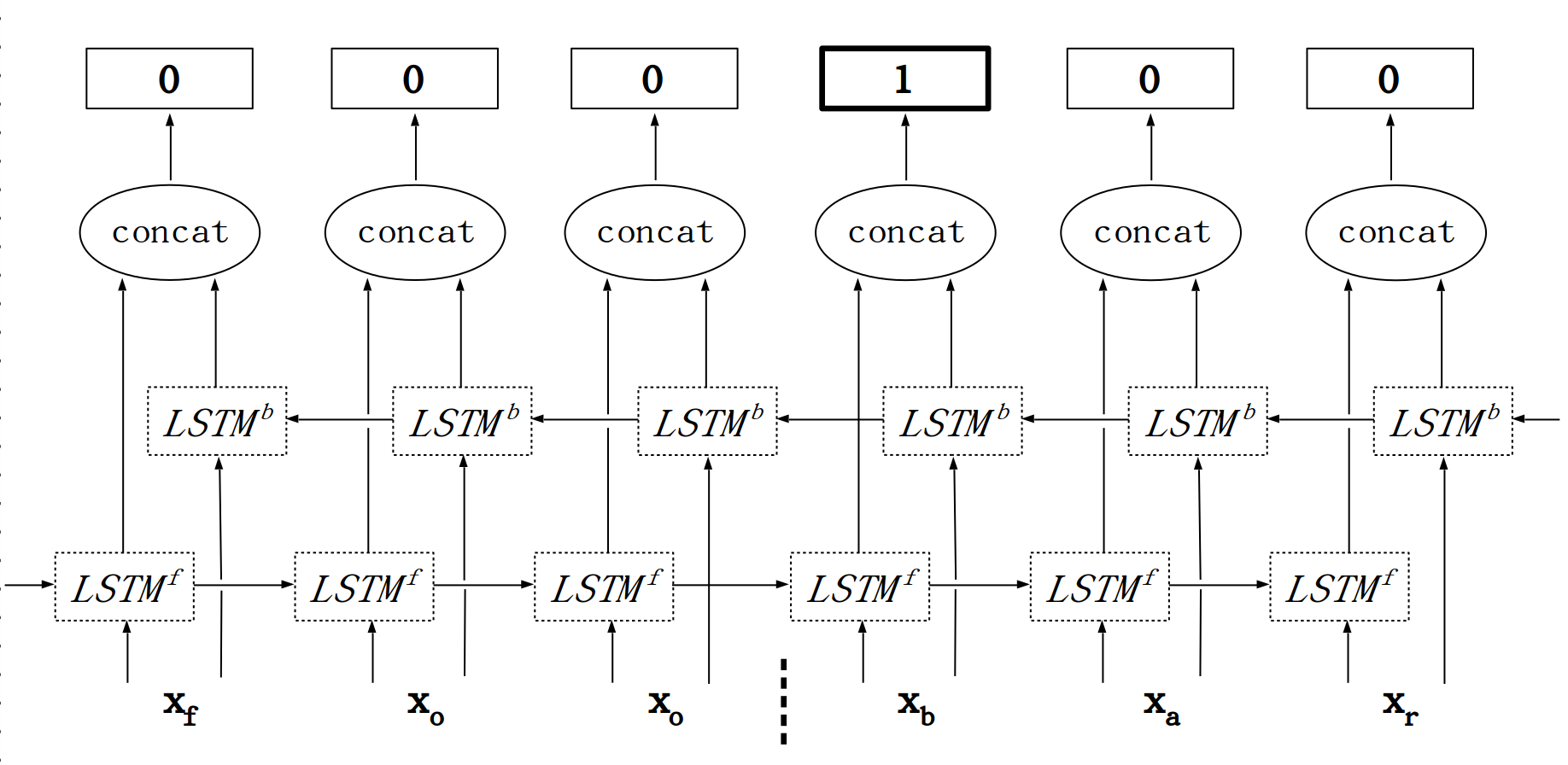}
\caption{BiLSTM network with one layer running on \texttt{foobar}. The vertical dashed line indicates the separation point.}
\label{bilstm}
\end{figure}

\vspace*{-0.5em}
Fig. \ref{bilstm} demonstrates the architecture of a BiLSTM network to split identifiers. It processes the characters of each identifier in both directions. The schema contains a single recurrent layer for simplicity, however, the real network is built with two stacked layers. The second recurrent layer is connected to the time-distributed dense layer with binary outputs and sigmoid activation. An output of \texttt{1} means the character is a split point and \texttt{0} it is not. Sigmoid activation was used instead of softmax because there can be more than one split point per identifier.

We trained our BiLSTM network on two NVIDIA GTX 1080 GPUs using \textbf{Keras} \cite{chollet2015keras} with a \textbf{Tensorflow} backend \cite{tensorflow}. It took approximately \bilstmHoursToRunEpochs hours to complete \bilstmNumEpochs epochs. Table \ref{params} lists the hyperparameters we chose using Hyperopt \cite{hyperopt}. The training curves are on Figure \ref{bilstmcurve}.

\newcolumntype{M}[1]{>{\raggedright}m{#1}}

\vspace*{-0.5em}
\begin{figure}[H]
\centering
\begin{minipage}{.5\textwidth}
\centering
\begin{tabular}{M{4cm} r}
\hline
RNN	sequence length & \identifierLengthLimit \tabularnewline
Layer sizes & \bilstmLayerSize, \bilstmLayerSize  \tabularnewline
Batch size & 512  \tabularnewline
Epochs & \bilstmNumEpochs  \tabularnewline
Optimizer & Adam \cite{adam}  \tabularnewline
Learning rate & 0.001  \tabularnewline
\hline
\end{tabular}
\renewcommand{\figurename}{Table}
\setcounter{figure}{\value{tablecounter}}
\caption{Network train parameters}
\renewcommand{\figurename}{Figure}
\refstepcounter{tablecounter}\label{params}
\end{minipage}%
\begin{minipage}{.5\textwidth}
\setcounter{figure}{\value{figure}}
\centering
\includegraphics[scale=0.3]{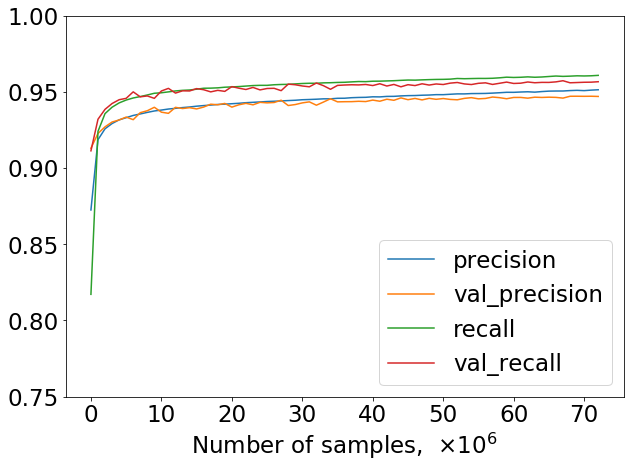}
\setcounter{figure}{\value{figurecounter}}
\captionof{figure}{Training curves for the BiLSTM}
\refstepcounter{figurecounter}
\label{bilstmcurve}
\end{minipage}
\end{figure}
\vspace*{-2em}

\section{Evaluation}

We divided the dataset into 80\% train and 20\% validation and calculated precision, recall and $F_1$ score for each of the models.
Precision is defined as the ratio of correct splitting predictions and the total number of predictions, recall as the ratio of correct splitting predictions and the ground truth number of splits, and $F_1$ score is the harmonic average of precision and recall. The results are shown on Fig. \ref{comparison} and Table \ref{eval}.
The worst models are clearly the statistical ones, however, the conjunction of character-level ML LMs achieved the highest precision among all with 96.6\%. Character-level CNN is close to the top, it has great evaluation speed and can be chosen if the run time is important. LSTM performed better than GRU and achieved the highest $F_1$ score with \bilstmPrecision\% precision and \bilstmRecall\% recall.

\vspace*{-0.1em}
\begin{figure}[H]
\centering
\begin{minipage}{.5\textwidth}
\centering
\footnotesize
\begin{tabular}{M{3cm} r r r}
Model & Precision & Recall & $F_1$ \tabularnewline
\hline
Char. ML LM $\rightarrow\lor\leftarrow$ & 0.563 & 0.936 & 0.703 \tabularnewline
Char. ML LM $\rightarrow\land\leftarrow$ & 0.966 & 0.573 & 0.719 \tabularnewline
Stat.\,dyn.\,prog.,\,Wiki & 0.741 & 0.912 & 0.818 \tabularnewline
Stat.\,dyn.\,prog.,\,Zipf & 0.937 & 0.783 & 0.853 \tabularnewline
Stat.\,dyn.\,prog.,\,posterior & 0.931 & 0.892 & 0.911 \tabularnewline
GBDT & 0.931 & 0.924 & 0.928 \tabularnewline
Char. CNN & 0.922 & 0.938 & 0.930 \tabularnewline
Char. BiGRU & 0.945 & 0.955 & 0.949 \tabularnewline
Char. BiLSTM & 0.947 & 0.958 & 0.952 \tabularnewline
\hline
\end{tabular}
\renewcommand{\figurename}{Table}
\setcounter{figure}{\value{tablecounter}}
\caption{Evaluation results}
\renewcommand{\figurename}{Figure}
\setcounter{figure}{\thefigure}
\refstepcounter{tablecounter}
\label{eval}
\end{minipage}%
\begin{minipage}{.5\textwidth}
\centering
\includegraphics[scale=0.3]{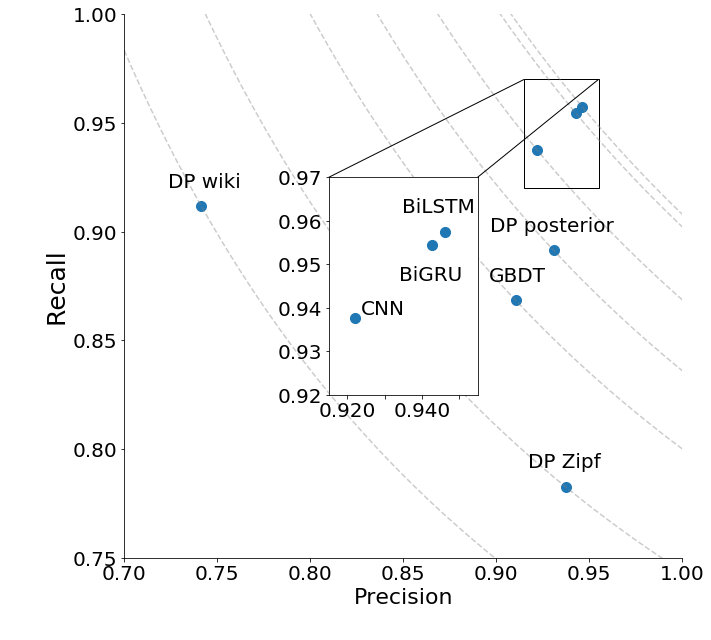}
\vspace*{-1em}
\setcounter{figure}{\value{figurecounter}}
\captionof{figure}{Model\,comparison,\,$F_1$\,isocurves\,are\,dashed}
\label{comparison}
\refstepcounter{figurecounter}
\end{minipage}
\end{figure}
\vspace*{-2em}

\section{Applications}
The presented identifier splitters reduce the number of unique subtokens by \ratioCoreVocabularyReduced\%. We ran the BiLSTM model on the subtoken sequences from the dataset and generated refined subtoken sequences. We then measured the new number of unique identifier parts, which was reduced from \ratioCoreVocabularyBefore to \ratioCoreVocabularyAfter. Samples of identifiers split by heuristics and our model are listed in appendix \ref{samples}.
Smaller vocabulary size leads to faster training of the upstream models such as identifier embeddings based on the structural co-occurrence scope as a context \cite{id2vec} or topic models of files and projects \cite{Markovtsev:2017}.

It is also possible to use our model to automatically split identifiers written in the same case without whitespace on the keyboard. This simplifies and speeds up typing the code provided by the number of splitting errors is low enough. Depending on the naming style, the described algorithm may save "Shift" or "Shift + Underscore" keystrokes. The reached quality metrics are good enough, our network makes an error with 50\% probability after $\log_{0.\bilstmPrecision}{0.5}\approx\bilstmIdentifiersToHalfError$ identifiers assuming that each identifier contains a single split point.

\section{Conclusion}
We created and published a dataset with \datasetRecordsAfterManualFiltering million distinct source code identifiers extracted from Public Git Archive, the largest one to date. We trained several machine learning models on that dataset and showed that the character-level bidirectional LSTM recurrent neural network (BiLSTM) performs best, reaching \bilstmPrecision\% precision and \bilstmRecall\% recall on the validation set. To our knowledge, it is the first time RNNs were applied to the source code identifier split problem. BiLSTM significantly (by \ratioCoreVocabularyTimes\ times) reduces the core vocabulary size in upstream problems and is good enough to improve the speed at which people write code.

\bibliographystyle{eptcs}
\bibliography{generic}

\begin{appendices}
\section{Identifier splitting algorithm, Python 3.4+} \label{tokenparser}
\begin{lstlisting}[language=Python]
NAME_BREAKUP_RE = re.compile(r"[ˆa-zA-Z]+")
min_split_length = 3

def split(token):
  token = token.strip()
  prev_p = [""]

  def ret(name):
    r = name.lower()
    if len(name) >= min_split_length:
      yield r
      if prev_p[0]:
        yield prev_p[0] + r
        prev_p[0] = ""
    else:
      prev_p[0] = r

  for part in NAME_BREAKUP_RE.split(token):
    if not part:
      continue
    prev = part[0]
    pos = 0
    for i in range(1, len(part)):
      this = part[i]
      if prev.islower() and this.isupper():
        yield from ret(part[pos:i])
        pos = i
      elif prev.isupper() and this.islower():
        if 0 < i - 1 - pos <= min_split_length:
          yield from ret(part[pos:i - 1])
          pos = i - 1
        elif i - 1 > pos:
          yield from ret(part[pos:i])
          pos = i
      prev = this
    last = part[pos:]
    if last:
      yield from ret(last)
\end{lstlisting}

\section{Examples of identifiers from the dataset processed by heuristics and the BiLSTM model}

\bgroup
\def\arraystretch{1.3}
\begin{table}[H]
\begin{center}
\centering
\footnotesize
\begin{tabular}{l l l}
\normalsize \textbf{Input identifier} & \normalsize \textbf{Output TokenParser} & \normalsize \textbf{Output BiLSTM} \tabularnewline
\hline
\texttt{OMX_BUFFERFLAG_CODECCONFIG}  & \texttt{omx bufferflag codecconfig} & \texttt{omx buffer flag codec config} \tabularnewline
\texttt{metamodelength}  & \texttt{metamodelength} & \texttt{meta mode length} \tabularnewline
\texttt{rESETTOUCHCONTROLS}  & \texttt{r esettouchcontrols} & \texttt{reset touch controls} \tabularnewline
\texttt{ID_REQUESTRESPONSE}  & \texttt{id requestresponse} & \texttt{id request response} \tabularnewline
\texttt{\%afterfor}  & \texttt{afterfor} & \texttt{after for} \tabularnewline
\texttt{simpleblogsearch}  & \texttt{simpleblogsearch} & \texttt{simple blog search} \tabularnewline
\texttt{namehash_from_uid}  & \texttt{namehash from uid} & \texttt{name hash from uid} \tabularnewline
\texttt{GPUSHADERDESC_GETCACHEID}  & \texttt{gpushaderdesc getcacheid} & \texttt{gpu shader desc get cache id} \tabularnewline
\texttt{oneditvaluesilence} & \texttt{oneditvaluesilence} & \texttt{on edit value silence} \tabularnewline
\texttt{XGMAC_TX_SENDAPPGOODPKTS} & \texttt{xgmac tx sendappgoodpkts} & \texttt{xgmac tx send app good pkts} \tabularnewline
\texttt{closenessthreshold}  & \texttt{closenessthreshold} & \texttt{closeness threshold} \tabularnewline
\texttt{test_writestartdocument}  & \texttt{test writestartdocument} & \texttt{test write start document} \tabularnewline
\texttt{dspacehash}  & \texttt{dspacehash} & \texttt{d space hash} \tabularnewline
\texttt{testfiledate}  & \texttt{testfiledate} & \texttt{test file date} \tabularnewline
\texttt{ASSOCSTR_SHELLEXTENSION}  & \texttt{assocstr shellextension} & \texttt{assoc str shell extension} \tabularnewline
\hline
\end{tabular}
\end{center}
\end{table}
\label{samples}
\egroup

\end{appendices}

\end{document}